\documentclass[preprint,12pt]{elsarticle}

\usepackage{algorithm,algcompatible,amsmath}
\algnewcommand\INPUT{\item[\textbf{Input:}]}%
\algnewcommand\OUTPUT{\item[\textbf{Output:}]}%

\usepackage{amssymb}
\usepackage{enumitem}
\usepackage{multirow}
\usepackage{color}
\usepackage[flushleft]{threeparttable}
\usepackage{hyperref}
\hypersetup{colorlinks=true,linkcolor=blue,anchorcolor=blue,citecolor=blue,filecolor=blue, urlcolor=blue,bookmarksnumbered=true,pdfstartview=FitB}
\setcitestyle{square}
\usepackage{makecell, caption}
\usepackage{amssymb,changes}

\journal{XXXX}
\biboptions{longnamesfirst,angle,semicolon}

\newcommand{\bd}{\mathbf{d}}

\begin{document}

\begin{frontmatter}



\title{Confirmatory Aspect-based Opinion Mining Processes}


\author[stat1]{Jongho Im \corref{cor1}}
\ead{ijh38@yonsei.ac.kr}

\author[busi3]{Taikgun Song}
\ead{Taikgun.Song@warrington.ufl.edu}

\author[busi1]{Youngsu Lee}
\ead{ylee54@csuchico.edu}

\author[busi2]{Jewoo Kim}
\ead{jjawoo@iastate.edu}

\cortext[cor1]{Corresponding author. Address: Yonsei University, Seoul, South Korea 03722. Tel: +82 2 2123 2539.}

\address[stat1]{Department of Applied Statistics, Yonsei University}
\address[busi3]{College of Business, University of Florida, Gainsville.}
\address[busi1]{Department of Finance and Marketing Department, California State University, Chico.}
\address[busi2]{Department of Apparel, Events, and Hospitality Management, Iowa State University.}

\begin{abstract}
A new opinion extraction method is proposed to summarize unstructured, user-generated content (i.e., online customer reviews) in the fixed topic domains. To differentiate the current approach from other opinion extraction approaches, which are often exposed to a sparsity problem and lack of sentiment scores, a confirmatory aspect-based opinion mining framework is introduced along with its practical algorithm called DiSSBUS. In this procedure, 1) each customer review is disintegrated into a set of clauses; 2) each clause is summarized to bi-terms-a topic word and an evaluation word-using a part-of-speech (POS) tagger; and 3) each bi-term is matched to a pre-specified topic relevant to a specific domain. The proposed processes have two primary advantages over existing methods: 1) they can decompose a single review into a set of bi-terms related to pre-specified topics in the domain of interest and, therefore, 2) allow identification of the reviewer's opinions on the topics via evaluation words within the set of bi-terms. The proposed aspect-based opinion mining is applied to customer reviews of restaurants in Hawaii obtained from TripAdvisor, and the empirical findings validate the effectiveness of the method.
\end{abstract}


\end{frontmatter}


\section{Introduction \label{sec1}}

User-generated contents (UGCs) available on social media and online communities are unlimitedly growing. Hence, the deluge of customer reviews and feedback became an important source for assessing customer satisfaction, product quality, and corporate competitiveness. A number of studies have demonstrated the usefulness of UGC reviews in understanding the determinants of customer experience and firm performance in business and the hospitality and tourism industries in particular \cite{Kim16, Viglia16, Xiang15}.

A variety of approaches are suggested to identify and assess the opinions of topics that customers discuss in online reviews. Early studies took a coarse-based approach to topic identification, in which all customer reviews are aggregated at a document level, and then topics are identified based on assumptions of probabilistic distribution of words  \cite{Blei03, Hofmann99}. Due to its aggregated form of data inputs, extracted words for a certain topic are often mixed such that unrelated words are grouped together to represent a topic. Therefore, it is hard to interpret and label the topic based on the extracted words. Due to this drawback from the aggregated form of data inputs, following studies adopted more fine-grained approaches, such as using a sentence as an input to identify topics and then extracting relatively consistent words for a topic \cite{Titov08, Brody10}. However, fine-grained approaches using sentences are also prone to have the data sparsity issues such that the sparse words appear concurrently in individual documents \cite{cheng14}. In addition to extracting relevant words for a topic, another area of research focuses on extracting sentiment words using linguistic knowledge \cite{socher13}. Though this approach is accessible to more general situations, the analysis may not be achievable without knowledge of the domain of interests.

To overcome those limitations, we propose a new technique that identifies the meaning structure of short text reviews based on topic matching that utilizes topic modeling and domain knowledge. The new method is distinct from other methods in several ways. First, the new method uses a clause as a unit of analysis instead of using a document or a sentence. A sentence-level probabilistic generative model may produce fine-grained topics compared to a document-level model. However, a sentence-level analysis may not be appropriate when one sentence includes multiple features and topics, and the opinion on each topic also may not be clear. Take a simple customer review, ``Food was good, but staff was rude," as an example. Two topics-food and service-are included, and two mixed sentiments-good and rude-are shown in one sentence. Furthermore, in the case of lengthy sentences, one needs to take additional computation works in identifying opinion sentences.

Second, topics related to a specific domain are generally specified in advance. Unlike other approaches that interpret and label extracted topics post hoc, the proposed method allows researchers or users to choose topics for the domain of interest before extracting opinions by using ones' domain knowledge. Then, the topics discussed in UGC reviews are identified by matching the extracted words with the pre-specified topics.  In the previous example sentence, one may be interested in not only the quality of food but also the quality of service. When staff is assigned to the service aspect, more targeted opinion extraction is possible. Therefore, we refer to our method as \textit{confirmatory aspect-based opinion mining}, which is analogous to confirmatory factor analysis in that the topics of interest are matched with certain words beforehand.

For the proposed method that identifies fine-grained topics discussed at a clause level and their sentiments from UGC, we introduce a series of processes: Disintegrating, Summarizing, Straining, Bagging, Upcycling, and Scoring (DiSSBUS). We develop this procedure by building upon an aspect-based opinion mining algorithm, which extracts aspects (topics) and opinions. Under this procedure, first, retrieved short texts (e.g., customer reviews) are disintegrated into a set of clauses instead of sentences. Second, the disintegrated clauses are summarized into a set of word pairs (i.e., bi-terms) using a parsing algorithm, such as Google SyntaxNet \cite{Syntaxnet16}. This step allows sets of bi-terms to maintain the original meanings of each review. Third, few specific bi-terms are strained from the set of bi-terms according to their relative frequencies in the second step and then classified as a representative expression for training. Fourth, the strained bi-terms are manually matched to fixed topics that are pre-specified by researchers using domain knowledge and external information. Fifth, using the matched bi-terms within each topic, unstrained bi-terms are also matched to the pre-specified topics, and therefore bags of bi-terms are expanded. Finally, the sentiment of the clustered bi-terms is scored, focusing on evaluation words.

Our aspect-based opinion mining approach has several advantages over existing methods. First, the proposed aspect-based opinion mining takes advantage of every single short text or short clause and, therefore, will significantly reduce information loss. Second, it integrates further reviewers' attitudes toward topics using sentiment analysis on evaluation words within the bi-terms set. Third, since topics are pre-specified, the proposed aspect-based opinion mining approach allows researchers and practitioners to focus on topics and areas of their interests.

This paper is organized as follows: in Section \ref{sec2}, we give a brief review of related works. The basic setup and terminology are introduced in the following Section \ref{sec3}, followed by detailed procedure in Section \ref{sec4}. Then, Section \ref{sec5} applies the proposed approach to online customer reviews of restaurants in Honolulu retrieved from TripAdvisor. Conclusion and implications of this study are discussed lastly in Section \ref{sec6}.

\section{Literature review \label{sec2}}
\subsection{Aspect-based opinion mining}

Our research question is how we can use unstructured text reviews to extract variables of interest for research when the variables cannot be directly obtained in a structured form. We address this question by summarizing text reviews into a vector of topics conserving their original linguistic meanings, semantics, and sentiment. As presented in Table \ref{tab1}, for example, short text reviews are summarized into ordered responses on the five pre-specified topics such that it takes three-level likert scale (1,2,3) if a pre-specified topic is stated within a review and zero otherwise:
\begin{table}[!h]
	\centering
	\caption{Illustrative example for structured data \label{tab1}}
	\begin{tabular}{ccccccc}
		\hline Text & Topic1 & Topic2 & Topic3 & Topic4 & Topic5 \\
		\hline
		1 & 1 & 0 & 1 & 2 & 1\\
		2 & 2 & 0 & 0 & 1 & 0\\
		3 & 3 & 2 & 2 & 2 & 3\\
		\hline
	\end{tabular}
\end{table}
The converted online reviews are useful in general statistical analyses, such as classification, regression, clustering, and factor analysis.

To achieve our research goal, which projects unstructured text data onto the pre-specified topic domains, we first consider aspect-based opinion mining originated from aspect-based summarization, which extracts aspect (topic) and opinion sentences and then identifies the sentiment orientation at sentence level \cite{hu04, scaffidi07, homoceanu11, eirinaki12}.  
Among similar approaches, Hu and Liu \cite{hu04} initially proposed aspect-based summarization which consists of three computational steps: 1) aspect identification, 2) opinion extraction, and 3) summary generation. According to the proposed approach, the product's aspects are extracted from the available reviews in the first step. Given the identified aspects, opinion sentences are extracted when the sentences are relevant to the features in the second step. At the final step, the selected opinion sentences are aggregated within each aspect to generate the final summary.

Although the aspect-based opinion mining methods look promising to analyze online reviews, they still have several limitations in application to the business domain. First, the current methods may not be applied to the cases where product or service aspects are not directly presented in the text. Regarding this issue, current research has focused on identifying aspects that include not only physical features but also abstract concepts \cite{taylor14,liu16}. Marrese-Taylor et al. \cite{taylor14} applied Hu and Liu \cite{hu04}'s algorithm to the tourism domain. Marrese-Taylor and colleagues considered customer reviews as a set of aspect expressions and then extracted opinion words for aspect identification. Liu et al. \cite{liu16} proposed a fine-grained aspect identification method by offering two unsupervised methods in rule set selection.

Second, a sentence-level analysis may not be appropriate when one sentence contains multiple aspects, and the reviewer’s opinions on the multiple aspects are different and even opposite. Furthermore, in the case of lengthy sentences, additional computation works are needed to identify opinion sentences. Third, sentences with no aspects and no opinion words result in missing values. To overcome these substantial problems, we propose a confirmatory aspect-based opinion mining that allows pre-specified aspects and decomposes a review as a set of clauses, not sentences.

\subsection{Related works}

The proposed algorithm is an extension of the aspect-based opinion mining algorithm that considers the analysis of text reviews at a more granular level. The key computational works are clause segmentation, clause summarization, semantic similarity measure, and sentiment lexicons. These related works are briefly introduced along with the basic idea of the proposed algorithm.

We use a clause as a lexicon unit because a clause includes an object and a predicate associated with a certain topic domain. Thus, \textit{clause segmentation} algorithm is required to split a sentence into a set of clauses. Clause segmentation is implemented by detecting a clause boundary, which is non-trivial and often ambiguous due to punctuation marks, conjunctions, and so on, as in sentence boundaries \cite{ram08}. The previous methods for clause boundary detection can be partitioned into rule based \cite{papageorgiou97, leffa98, gadde10}, machine learning-based \cite{sang01, molina02}, and hybrid approaches \cite{ram08, orasan00}.The rule-based algorithm parses an input sentence and then finds a clause boundary within each clause type specified in advance, while the boundary classifier is trained using a statistical model in the machine learning-based approach. Motivated by \textit{sentence compression}  that considers a combination of several meaningful words as a ``scaled down version of the text summarization" \cite{knight02}, the proposed algorithm summarizes each segmented clause into a set of words. Sentence compression algorithms \cite{knight02,mcdonald06, galley07, clarke08, berg11, filippova15} determine whether each word token in the input sentence should be kept or dropped. Thus, as the result of sentence compression, a simpler sentence is produced after deleting less informative words.

The proposed algorithm, presented in this paper, requires semantic similarity scores between words. For example, it is required to determine if ``delicious food" and ``amazing steak" should be clustered into the same topic. There are two types of semantic similarity measures: corpus-based and knowledge-based \cite{mihalcea06}. In the corpus-based measures, the degree of similarity between words is identified using information derived from the corpus. The algorithms are usually based on the statistical analysis of word usage to capture the occurrences or co-occurrences of words within the text. The algorithms often cannot be expressed with a simpler equation or format. Metrics such as point-wise mutual information \cite{turney01}, latent semantic analysis \cite{landauer98}, and topic modeling such as LDA \cite{Blei03} are typical corpus-based measures. Compared to the corpus-based approaches, knowledge-based measures utilize additional information obtained from semantic networks to quantify the degree to which two words are semantically associated \cite{budanitsky01}. These measures mainly use a lexical database, such as WordNet \cite{miller95, book_fellbaum98}. The present paper proposes a modified knowledge-based similarity measure utilizing website such as thesaurus.com.  

Numerous methods have been proposed for sentiment analysis or opinion mining. As a remarkable reference, Ravi and Ravi \cite{rabi15} explored these numerous sentiment analysis methods at word, clause, sentence, and document levels. For sentiment classification, polarity determination algorithms generate ordered scores of topics. The polarity determination is classified into machine learning-based approaches \cite{pang02,dang10, zhang11} and lexicon-based approaches \cite{turney02,kamps02,nasukawa03, benamara07}. In the machine learning-based approaches, the sentiment level is identified with statistical methods, such as Naive Bayes (NB), Support Vector Machine (SVM), and Neural Network (NN). The lexicon-based approaches employ a dictionary of sentiment words called \textit{sentiment lexicon} \cite{book_liu15}. For example, each word is assigned a sentiment score based on a predetermined sentiment dictionary or WordNet. In this paper, we use a lexicon-based scoring approach to avoid the sparsity problem and simplify our proposed algorithm. The details are introduced in the next section.

\section{Confirmatory aspect-based opinion mining \label{sec3}}

In this section, we discuss the theoretical background of how the proposed method mimics the maximization of likelihood approach. In principle, the method converts the full joint likelihood of reviews as the product of the conditional probability of topics given a clause and the marginal probability of occurrence the clause. We first define terminology with notation and introduce key assumptions to discuss our aspect-based opinion mining method. In addition to basic definitions often used in text analysis, such as word, document, topic, and corpus, we newly define clause, object word, and evaluation word, which distinguish our aspect-based topic matching approach from the conventional methods. 

\begin{enumerate}[label=\textbf{(N\arabic*)},ref=(N\arabic*)]
\item \label{ns1} A {\it word} denoted by $w$ is the basic unit of discrete text data.
\item \label{ns2} A {\it clause} denoted by $c$ is a sequence of words that includes both a subject and a predicate.
\item \label{ns3} A {\it document} is a sequence of words or a set of clauses denoted by $\bd=(w_{1},\dots, w_{m})$ or $\bd=(c_1,\dots, c_s)$, where $w_{j}$ is the $j$-th word, $j=1,\dots,m$ and $c_k$ is the $k$-th clause, $k=1,\dots,s$.
\item \label{ns4} A {\it topic} denoted by $t$ is a subject of discussion or review.
\item \label{ns5} A {\it corpus} is a collection of $n$ documents denoted by $V=\{\bd_1,\dots, \bd_n\}$.
\item \label{ns6} An {\it object word} denoted by $o$ is a noun or verb used for subject, object or complement in a clause.
\item \label{ns7} An {\it evaluation word} denoted by $e$ is a set of adjective, adverb or verb that describes or predicates an object word within a clause.
\end{enumerate}	

The main idea of the proposed algorithm is that any document can be summarized with a set of bi-terms while the original meaning of the documents is preserved as far as possible. A document consists of distinct clauses, and the clauses can be projected to a space that is spanned by a set of bi-terms whose textual meanings are essentially the same as the clauses. For a given $k$-th clause at the $i$-th document $c_{ik}$, we can summarize the clause with a bi-term $b_{ik}=(o_{ik},e_{ik})$, where $e_{ik}$ is a key evaluation word in the clause $c_{ik}$ and $o_{ik}$ is an object word corresponding to the $e_{ik}$. Note that $o_{ik}$ can be a set of topic words, but we simply assume a single word for brevity. For example, in the online restaurant review, ``The food was amazing and the service was great",
there are two clauses: ``the food was amazing" and ``the service was great". Both clauses can be reduced to two bi-terms (food, amazing) and (service, great). Since the proposed computation algorithms are implemented on bi-terms, the performance of the proposed method depends on whether or not each clause can be properly disintegrated to a bi-term.

To discuss our aspect-based opinion mining approach in a statistical framework, we first consider a full joint likelihood:
\begin{equation}\label{eq:fulllike}
p(\bd_1, \dots, \bd_n).
\end{equation}
It is very difficult to handle the likelihood directly (\ref{eq:fulllike}) because it is essentially the full joint likelihood of a potentially large number of discrete words. To construct feasible topic modeling, we assume that  

\begin{enumerate}[label=\textbf{(A\arabic*)},ref=(A\arabic*)]
\item \label{as1} Documents are independent each other. That is, the full likelihood in (\ref{eq:fulllike}) can be written again such that 
$$p(\bd_1, \dots, \bd_n )=\prod_{i=1}^n p(\bd_i).$$

\item \label{as2}  Each clause can be approximated a bi-term that consists of an objective word $o$ and an evaluation word $e$ and then we can write
\begin{eqnarray*}
p(\bd_i)&=&p(w_{i1},\dots, w_{im} ) \\
        &=&p(c_{i1},\dots, c_{is} )  \\
        &=&p\{(o_{i1},e_{i1}),\dots, (o_{is},e_{is})\}. 
\end{eqnarray*}

\item \label{as3} Given a $p$-dimensional topics, a bi-term $b_{ik}=(o_{ik},e_{ik})$ is associated with exactly one topic and let $t_{ik}$ be the matched topic for $b_{ik}$. 

\item \label{as4} Clauses within a document are conditionally independent given a topic and then the marginal likelihood $p(\bd_i)$ can be defined by
\begin{eqnarray*}
p(\bd_i )
&=&\sum_{l=1}^p \prod_{k=1}^s p(o_{ik},e_{ik}\mid t_l)p(t_l), 
\end{eqnarray*}                                                              
where $t_l$ $(l=1,\dots, p)$ denotes the $l$-th topic.                                           
\end{enumerate}

Assumption \ref{as1} is a natural condition. If a hidden network of reviews can be identified, one may allow a dependency structure between them. In general, however, it is reasonable to assume that all reviewers are independent of each other. This assumption indicates that each review has its own topic mechanism, and topics for each review can be independently determined regardless of other reviews

Assumption \ref{as2} states a dimension reduction. Instead of incorporating a unigram or multigram condition, we apply aspect-based opinion mining through NLP. Each review consists of several sentences, and each sentence consists of clauses. Each clause is summarized in two words called bi-terms. Eventually, a sentence is represented by a set of bi-terms.  

\begin{figure}[h]
	\begin{center}
		\includegraphics[width=20em]{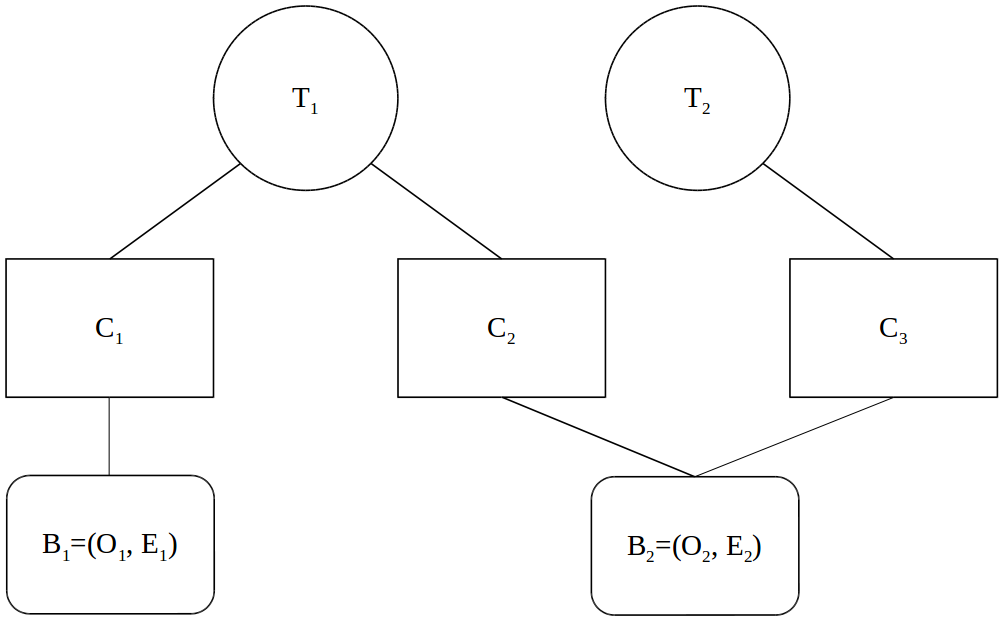}
	\end{center}
	\caption{Matching rule for multiply connected bi-terms \label{fig1}}
\end{figure}

Assumption \ref{as3} implies that a clause is associated with a single topic. Referring to the previous example, ``food" is the topic for a clause ``the food was amazing" and ``service" is is the topic for the clause ``the service was great".However, this assumption does not imply that each bi-term should be associated with a single topic because different clauses can be summarized into the same bi-terms. Figure \ref{fig1} presents a case where clauses are connected to a single topic, and a bi-term is linked to two topics through two different clauses.

\begin{figure}[ht]
	\begin{center}
		\includegraphics[width=30em]{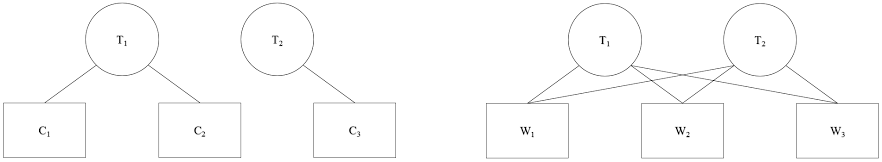}
	\end{center}
	\caption{Left Panel: Proposed aspect-based opinion mining; Right Panel: Classical LDA \label{fig2}}
\end{figure}

Figure \ref{fig2} presents the difference between the proposed aspect-based opinion mining with other approaches, such as Latent Dirichlet allocation (LDA). The proposed algorithm assigns a single topic on each clause, while other approaches estimate all associated probabilities between topics and words. Therefore, under the framework of the proposed algorithm, a clause can be partially correlated with other clauses through a matched topic, but they are independent given this topic by the assumption \ref{as4}. 

Based on \ref{as1}-\ref{as4}, the full joint likelihood can be expressed as a p-dimensional vector of topics,
\begin{eqnarray}
p(\bd_1, \dots, \bd_n) 
&=&\prod_{i=1}^n\sum_{l=1}^p \prod_{k=1}^s p(c_{ik}\mid t_l)p(t_l)\notag\\
&=&\prod_{i=1}^n\sum_{l=1}^p \prod_{k=1}^s p(t_l \mid c_{ik})p(c_{ik})\label{eq:clike}\\
&\simeq&\prod_{i=1}^n\sum_{l=1}^p \prod_{k=1}^s p(t_{l}\mid o_{ik}, e_{ik})p(o_{ik}, e_{ik})\label{eq:like}
\end{eqnarray}
The likelihood (\ref{eq:clike}) indicates that a topic modeling is converted to estimate the conditional probability of a topic in a given clause. This clause-based estimation can be achieved utilizing a text clustering algorithm with supervised or unsupervised learning classification. However, because a knowledge-based semantic measure is often more applicable to short text reviews, we use the likelihood (\ref{eq:like}) that is proportional to the likelihood (\ref{eq:clike}). The likelihood (\ref{eq:like}) can be maximized when topics are assigned to each clause so that it has the highest probability of a topic conditional upon the given bi-term. Thus, in this paper, we propose a new aspect-based opinion mining algorithm that tries to maximize the conditional probabilities of $p(t_{l}\mid o_{ik}, e_{ik})$.

\section{Confirmatory aspect-based opinion mining processes \label{sec4}}

We propose a series of algorithmic approaches to achieve confirmatory topic match. The process can be referred to in the abbreviated form, DiSSBUS, which stands for disintegrating, summarizing, bagging, upcycling, and scoring algorithms. The proposed process produces a set of topics with sentiment scores for each review. The details are given below:

\begin{enumerate}[label=(\textbf{S\arabic*}),ref=(S\arabic*)]
\item \label{st1} \textbf{Disintegrating:} Compress each clause to a bi-term with negation words if applicable.

\item \label{st2} \textbf{Summarizing:} Summarize each clause to a bi-term with negation words if applicable.

\item \label{st3} \textbf{Straining:} From the collection of bi-terms, extract specific bi-terms that are classified as a common expression, which is defined by being frequently and jointly used within clauses. Here, strained bi-terms can be viewed as a training sample and unstrained bi-terms as a testing sample in a machine-learning framework.

\item \label{st4} \textbf{Bagging:} Match the selected bi-terms to a pre-specified topic. This procedure is manually implemented if there is no external database.

\item \label{st5} \textbf{Upcycling:} Topics are selectively assigned on unstrained bi-terms according to the matching results of the strained bi-terms. 

\item \label{st6} \textbf{Scoring:} Generate sentiment scores based on sentiment lexicons. 
\end{enumerate}

\subsection{Disintegrating}

Using clauses has several advantages over other approaches that utilize full sets of documents or sentence in consumer reviews. First, the aspect-based opinion mining is more effective at a clause level than a sentence level because sentences are more likely to include multiple semantic meanings. For example, food and service are two different topics and can be included in one sentence, as shown in Section \ref{sec3}. However, separating food and service represents the topics in the review better. Thus, separated clauses is beneficial not only for dis-ambiguity but also interpretation and computational efficiency. Second, using clauses can improve the quality of parsing. For sentences to be converted into meaningful clauses, long text reviews need to be parsed using a parsing algorithm such as Google SyntaxNet \cite{Syntaxnet16} or NLP parser \cite{bird09}. Because the accuracy of parsing results depends on the complexity of texts \cite{delisle95}, clauses in simple forms are necessary. Furthermore, simple clauses consisting of a bi-term can be easily matched to pre-specified topics (aspects). 

\textit{Disintegrating} may not be easily implemented due to the ambiguity of punctuation marks. For example, the ``period" not only indicates the end of a sentence but also shows decimals in a number or separates letters in abbreviated words or acronyms. Also, the fact that online reviews often violate English grammar and spelling makes it difficult to categorize the use of a period without understanding the context. Abbreviations or acronyms can also be problematic because those terms often accompany ``period". Another issue with clause segmentation arises when a comma is used as a conjunction to connect two clauses. For example, in the sentence ``Before I ate dinner, I took a bus tour," a comma is used to connect two clauses ``Before I ate dinner" and ``I took a bus tour" into a sentence. However, not all commas and periods can be used as an indicator to separate documents; they could rather be used to list a series of items. For example, one should not separate a document such as ``I like to eat eggs, bacon, and pancake as breakfast" just because it has commas.

In order to handle these issues, we suggest the rule-based clause segmentation algorithm summarized in Algorithm \ref{algo:segmentation}. Using this algorithm, first, we identify the location of punctuation marks and conjunctions. Second, we test if the part-of-speech (POS) tags are similar to tags before punctuation marks or conjunctions. If a comma is used to separate two items, then the words before and after the comma will have the same POS tags. For example, in the sentence ``wow the food was simple, fresh, and delicious," commas were used simply to list words, not to connect clauses. Since a parser will tag ``simple", ``fresh", and ``delicious" as all adjectives, one could successfully identify the use of commas in this example. All punctuation marks and conjunctions other than the ones being used to list items will be considered as clause identifiers. Third, when clause identifiers are applied to the input sentence, multiple clauses will include subjects and predicates and phrases as word chunks.

\begin{algorithm}
\caption{Clause Segmentation}\label{algo:segmentation}
\begin{algorithmic}[1]
\REQUIRE parser
\INPUT  sentence or a vector of word sequence
\OUTPUT clauses or word chucks
\STATE parse input sentence.
\STATE enumerate punctuation marks and conjunctions in a set $B$.
\STATE \textbf{for} $b \in B$ \textbf{do}
\STATE  \quad test if $b$ is clause identifier.
\STATE \textbf{End for}
\end{algorithmic}
\end{algorithm}

\subsection{Summarizing}

The DiSSBUS algorithm is designed to simplify summarization, clustering, and scoring procedures. To achieve this goal, each clause is compressed into a bi-term. Based on parsing and tagging algorithms, aspect-based summarization extracts our target bi-terms from input clauses. The relationship of the selected bi-term in a clause is strongly connected with its POS. For example, when the object word is a noun, then the evaluation word is likely an adjective. Therefore, assigning POS tags for individual components in a clause vastly reduces the number of bi-term combinations. 

In the parsing and tagging results, there are five major combination types: (Noun, Adjective), (Noun, Verb), (Noun, Adverb), (Verb, Adjective), and (Verb, Adverb). Other combinations are not helpful for capturing the core meaning of input clauses. For example, (food, good) is extracted as the selected bi-term from an input clause, ``the food was very good". In practice, it is possible to have multiple bi-terms in a clause. One of those bi-terms keeps the primary semantic meaning, but others play as predicative for the main bi-term. However, we cannot automatically specify the main bi-term at this step. Redundant bi-terms are removed during the straining, bagging, and upcycling procedures.

\subsection{Straining}

In the straining step, bi-terms are shrunk to a smaller set of bi-terms called common expressions. Given a frequency tuning parameter $C$, common expressions are selected using the following straining algorithm:

\begin{algorithm}[!h]
\caption{Straining Algorithm}\label{algo:straining}
\begin{algorithmic}[1]
\INPUT  bag of bi-terms $B$, cut-pint $C$.
\OUTPUT bag of common expressions $B_C$.
\STATE  set $B_C=\emptyset$
\STATE \textbf{for} $b \in B$ \textbf{do}
\STATE \quad \textbf{if} $\mid b \mid \ge C$ \textbf{then}
\STATE \quad \quad $B_C \leftarrow B_C \cup b$
\STATE \quad \textbf{End if}
\STATE \textbf{End for}
\end{algorithmic}
\end{algorithm}

The size of the common expression can be adjusted by the tuning parameter $C$. If the parameter $C$ is relatively larger, the smaller set of bi-terms is extracted. Then, less manual work is required to topic match between the bi-terms and pre-specified topics. However, the overall quality of topic matching will decrease due to the deficiency of reference topic matching results obtained from small common expressions.

\subsection{Bagging}

For topic matching, a manual bagging (labeling) procedure is incorporated to create a topic classification rule. Using the following steps, the topics are matched with the selected common bi-terms:

\begin{enumerate}
	\item[(a)] Given a common expression bi-term, clauses containing the bi-term were extracted.
	\item[(b)] Clauses were manually read, and then the bi-term was classified to a topic bag that best described the clauses.
	\item[(c)] If there was no applicable topic or the bi-term was used over topics, the bi-term was discarded.
\end{enumerate}

The procedure (b) is constructed to have disjoint sets of bi-terms as the default setting. Although it is possible to have multiple bags at a clause level, as presented in Figure \ref{fig1}, we enforce a single topic on each bi-term using a \textit{softmax rule}.

\subsection{Upcycling}

The initial bag-of-topics is extended through the upcycling process. The matching results of common expressions are applied to the uncommon expressions (unstrained bi-terms) using a semantic similarity measure. A new topic is assigned to parts of the unstrained bi-terms using the softmax rule, and the unassigned bi-terms and clauses are finally ignored in the topic matching and scoring. This process is named by upcycling in the sense that some unstrained bi-terms are used for extending the bags of bi-terms and can be understood as the prediction or imputation for the unlabeled bi-terms. 

To implement the upcycling, we first define semantic similarity scores $s_{ij(k)}$ between the bi-term $i$ in the uncommon expressions and the $k$-th bag-of-biterms,
\begin{equation}\label{distm}
s_{ik}=\sum_{j=1}^{n_k}w_k\{1-d(x_{i,k}, x_{j,k})\},
\end{equation}
where $x_i$ denotes noun or verb in bi-term $i$, $n_k$ is the size of bag-of-biterms for topic $k$, $w_k$ is the topic weights, and $d(x_{i,k}, x_{j,k})$ denotes the semantic distance between nouns or verbs. The topic weight $w_k$ is associated with accuracy on imbalance data and can be understood as a learning rate in views of machine learning sense. 
However, in this paper, we set the equal value on all topics for simplicity. When nouns or verbs in two bi-terms are simultaneously included in a book of synonyms (e.g., thesaurus) dictionary  or a higher level of word classification category, the semantic distance $d$ has zero value. For example, both ``amazing steak" and ``disgusting hamburger" have zero values compared to the a bi-term ``great food" within the bag of ``food", because steak and hamburger can be clustered into the same food category. WordNet, a lexical database an online dictionary, and similar databases sentiment dictionaries can be used to calculate semantic similarity measures.

Once semantic similarity scores are computed for all possible topics, a topic associated with the highest score is assigned to the bi-term according to the softmax rule. During the upcycling procedure, we update the bag-of-topics to improve matching accuracy. If a newly assigned bi-term has the dominant score in a certain topic bag and is frequently used in a corpus, then we extend the matched topic bag to include the new bi-term. The score criterion  $(C_1)$ and relative frequency criterion $(C_2)$ are tuning parameters that control the size of the topic bags. The updating algorithm is described in Algorithm \ref{algo:bagging}.

\begin{algorithm}[!h]
\caption{Bagging Algorithm}\label{algo:bagging}
\begin{algorithmic}[1]
\REQUIRE WordNet or similar sentiment dictionary
\INPUT  Unstrained bi-terms $B$ and bag-of-topics $T=(T_1,\dots, T_K)$, criterion $C_1$ and $C_2$.
\OUTPUT Extended bag-of-topics $T=(T_1,\dots, T_K)$
\STATE \textbf{for} $b_i \in B$ \textbf{do}
\STATE \quad Compute similarity measure $s_{ij}$ to topic $j$
\STATE \quad \textbf{if} $s_{ij} \ge C_1$ and $n(b_i) \ge C_2$ \textbf{then}
\STATE \quad \quad $T_j \leftarrow T_j \cup b_i$
\STATE \quad \textbf{End if}
\STATE \textbf{End for}
\end{algorithmic}
\end{algorithm}

In the scoring calculation, we consider two options: \textit{full comparison} and \textit{fractional comparison}. In full comparison, the scores are computed over all the bi-terms in each topic bag. For fractional comparison, we first set a total comparison size $M$ and then allocate sample size $M_k$ on the topic domain that is proportional to the size of bags, where
$$M_k \propto M\frac{n_k}{\sum_{k=1}^K n_k}.$$
After obtaining a set of $M_k$, then randomly selected bi-terms with size of $M_K$ within the $k$-th bag-of-biterms. Finally, The scores are finally computed on the selected $M$ bi-terms.

As a result of the bagging and upcycling procedures, we have a number of bi-term bags that correspond to the size of the topics. These bags-of-biterms can serve as a baseline reference for similar reviews. For example, topic-matched bi-terms from TripAdvisor can be helpful to implement topic matching for another set of bi-terms derived from Yelp.

\subsection{Scoring}

The advantage of summarizing each clause into a topic-related objective word and its opinion word (evaluation word) naturally leads to sentiment analysis or scoring. For this process, a data set containing information of sentiment lexicons is required. The sentiment score for each (evaluation) word in a bi-term is obtained using the sentiment database. The scoring of a bi-term is then extended to all clauses that consist of the bi-term. For clauses containing negation words, such as ``not", ``neither", and ``don’t", the scores are given the opposite value of the bi-term score. For example, two clauses-``I do not like the food here" and ``I like the food here"-that have the same bi-term (like, food) will have opposite values based on the score of (like, food). The scores of the clauses are averaged within the same topic domain in each review. For this sentiment lexicon approach, we can also use WordNet or other sentiment databases as used in the upcycling step.

\section{Empirical analysis \label{sec5}}

We collect 6,983 TripAdviser reviews from 677 Hawaii restaurants using the program language \textbf{R}. The data set includes the restaurant name, overall rating, review title, customer review, and the date visited.  \autoref{tab:full_data} lists an example of the selected columns from the collected data set.

\begin{table}[ht]
\centering
\caption{First five rows of the data set with selected columns} 
\label{tab:full_data}
\begin{tabular}{llllc}
\hline
Title      & Review & Date & Rating \\ 
\hline
Fantastic!!     & great taste...   & 2016-04-18 &   5 \\ 
Went here...    & this place...    & 2016-04-12 &   5 \\ 
I'm dreaming... & this is hands... & 2016-04-11 &   5 \\ 
Amazing Poke!   & so glad we...    & 2016-04-11 &   5 \\ 
Great Poke      & just a small...  & 2016-04-09 &   4 \\ 
\hline
\end{tabular}
\end{table}

\begin{figure}[!h]
	\begin{center}
		\includegraphics[width=35em]{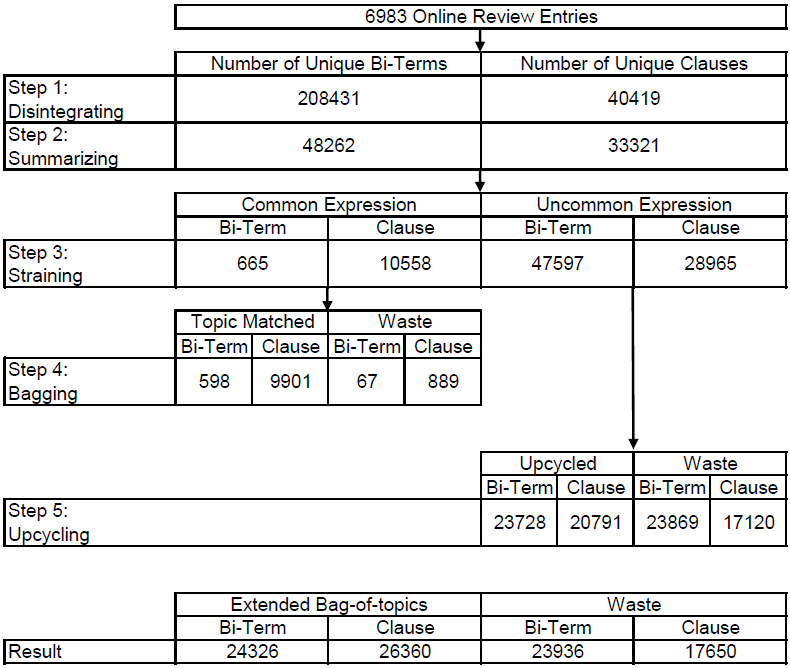}
	\end{center}
	\caption{Overall scheme of the DiSSBUS algorithm. \label{DiSSBUS}}
\end{figure}

Figure \ref{DiSSBUS} presents an overall scheme of the DiSSBUS algorithm and its corresponding output for each step of the TripAdvisor example. Each review was first disintegrated and then summarized into 48,262 bi-terms (33,321 corresponding clauses). Commonly expressed bi-terms were filtered out during the straining step, and 665 common expressions were manually assigned to one of eight possible topics: Food (F), Service (S), Price (P), Atmosphere(A), Environment (E), Revisit (Rv), Recommendation (Re), and Other (O). Borrowing from SERVQUAL, which was originally developed by Parasuraman et al. \cite{papageorgiou97}, we chose these pre-specified topics because customers are known to use these criteria when they assess service firms, particularly restaurants \cite{Lee95,Namkung08}. The bag-of-biterms was then extended using the upcycling step. Among the 48,262 bi-terms, the algorithm matched 24,326 bi-terms to eight main topics and classified the remains as irrelevant bi-terms in the pre-specified topic domains. The detailed step-by-step procedure is described below.

Each review was represented as a set of clauses or phrases using the proposed clause segmentation algorithm. 
\autoref{tab:disintegrating} shows an example of splitting an input sentence presented below: 

\bigskip

\textit{great taste, simple dish. never tried poke before. wow. go for the spicy dish. excellent. don't be discouraged by the way it looks from outside, it truly is a hole in the wall but, well worth it. plan on going again before we go back home.}

\bigskip

\begin{table}[!h]
\caption{An illustrative example of Disintegration} 
\label{tab:disintegrating}
\centering
\begin{tabular}{cl}
\hline
ID  & Clause \\ 
\hline
1 & great taste, simple dish. \\ 
2 & never tried poke before. \\ 
3 & wow. \\ 
4 & go for the spicy dish. \\ 
5 & excellent. \\ 
6 & don't be discouraged by the way it looks from outside,\\
  & it truly is a hole in the wall \\ 
7 & but, well worth it. \\ 
8 & plan on going again before we go back home. \\ 
\hline
\end{tabular}
\end{table}

Google SyntaxNet \cite{Syntaxnet16} was applied to each clause to tag the POS. Since the parsing results were provided in shape of a tree structure with identified root words, the bi-terms are also automatically arranged based on the Google SyntaxNet’s parsing results. Parsing results according to the example review above are presented in \autoref{tab:disintegrating tree}. 
Each row contains two words denoted by ``Word1" and ``Word2" with their specific POS information respectively. 

\begin{table}[!h]
\centering
{\footnotesize
\caption{An illustrative of parsing results using Google SyntaxNet} 
\label{tab:disintegrating tree}
\begin{tabular}{cllllll}
\hline
ID  & Original Clause & Tree & Word1 & POS1 & Word2 & POS2 \\ 
\hline
1-1 & great taste, simple dish. &  +-- great JJ amod      & great  & JJ & taste & NN \\ 
1-2 & great taste, simple dish. &  +-- , , punct          & ,      & ,  & taste & NN \\ 
1-3 & great taste, simple dish. &  +-- dish NN appos      & dish   & NN & taste & NN \\ 
1-4 & great taste, simple dish. &  $|$   +-- simple JJ amod & simple & JJ & dish   & NN \\ 
1-5 & great taste, simple dish. &  +-- . . punct          & .      & .  & taste & NN \\ 
2-1 & never tried poke before.  &  +-- never RB neg     & never  & RB & tried & VBD \\ 
2-2 & never tried poke before.  &  +-- poke NN dobj     & poke   & NN & tried & VBD \\ 
2-3 & never tried poke before.  &  +-- before RB advmod & before & RB & tried & VBD \\ 
2-4 & never tried poke before.  &  +-- . . punct        & .      & .  & tried & VBD \\ 
3   & wow.                      &  +-- . . punct & . & . & wow & UH \\ 
4-1 & go for the spicy dish.    &  +-- for IN prep         & for   & IN & go & VB \\ 
4-2 & go for the spicy dish.    &  $|$   +-- dish NN pobj    & dish  & NN & for   & IN \\ 
4-3 & go for the spicy dish.    &  $|$       +-- the DT det  & the   & DT & dish  & NN \\ 
4-4 & go for the spicy dish.    &  $|$       +-- spicy NN nn & spicy & NN & dish  & NN \\ 
4-5 & go for the spicy dish.    &  +-- . . punct           & .     & .  & go & VB \\ 
\hline
\end{tabular}
}
\end{table}

A total of 208,431 bi-term pairs were generated from combinations of parsing outputs. These initial pairs contain function words (determiners, interjections, conjunctions, prepositions,  \textit{et cetera}) that do not hold any lexical meaning. Therefore, it is logical to only focus on non-function words that include nouns, verbs, adjectives, and adverbs. Hence, the summarizing step was introduced to effectively decrease the size of the bi-term set by capturing the following bi-term combinations: (Noun, Adjective), (Noun, Verb), (Noun, Adverb), (Verb, Adjective) and (Verb, Adverb). As a result, 48,262 unique bi-term pairs with its POS tags were considered after the summarization process. As illustrated in Figure \ref{DiSSBUS}, each clause contains approximately 1.6 bi-terms on average after the summarizing step. Though this may contradict our assumption in \ref{as2}, slightly more than one single bi-term is acceptable considering that two types of bi-terms may exist in a clause: a bi-term that has own meaning, and a bi-term that predicates other bi-terms. 

Before conducting the straining step, a stemming procedure is implemented to enhance the performance of the matching outcome. For example, words such as ``stems," ``stemmer", ``stemming", and ``stemmed" are all converted to the a root word ``stem". After the straining step, a total of 665 common expression bi-terms were selected with a cut-point tuning parameter C=8. Some examples are presented in  \autoref{tab:CE}. 

\begin{table}[h]
\centering
\caption{Top 10 common expressions: noun (NN), verb(VB), adjective (ADJ), adverb (RB) \label{tab:CE}} 
\begin{tabular}{clclcc}
		\hline
	    Rank & Word 1 & POS 1 & Word 2 & POS 2 & Count \\ 
		\hline
		1 & food     & NN & good    & ADJ & 415 \\ 
		2 & food     & NN & great   & ADJ & 215 \\ 
		3 & servic   & NN & good    & ADJ & 192 \\ 
		4 & staff    & NN & friend  & ADJ & 168 \\
		5 & recommend& VB & highly  & RB  & 102 \\ 
		6 & place    & NN & great   & ADJ & 159 \\ 
		7 & go       & VB & back    & RB  & 153 \\
		8 & price    & NN & reason  & ADJ & 150 \\ 
		9 & servic   & NN & great   & ADJ & 149 \\ 
		10 & servic  & NN & friend  & ADJ & 109 \\
		\hline
\end{tabular}	
\end{table}

Each strained bi-term is matched to one of eight topics. All clauses containing each input bi-term were extracted and manually evaluated. In most cases, as illustrated in Table 6, commonly expressed bi-terms predominantly match one topic, which establishes a one-to-one relationship between a bi-term and a single topic. The top 10 common expressions with their assigned topics are provided in \autoref{tab:bagging}.

\begin{table}[!h]
\centering
\caption{Example clauses summarized to (well, done) \label{tab:toppic dim}} 
\begin{tabular}{l}
\hline
Original Clause \\ 
\hline
well done karai crab.\\ 
sometimes, i find the steak too well done\\ 
well done to the staff. \\ 
the garlic ahi was done well\\ 
flavors were really done well. \\
\hline
\end{tabular}
\end{table}

\begin{table}[ht]
\centering
\caption{Example common expressions with assigned topic \label{tab:bagging}} 
\begin{tabular}{rllllrl}
		\hline
		& Word 1 & POS 1 & Word 2 & POS 2 & Count & Topic \\ 
		\hline
		1 & food      & NN & good    & ADJ & 415 &  F \\ 
		2 & food      & NN & great   & ADJ & 215 & F \\ 
		3 & servic    & NN & good    & ADJ & 192 & S \\ 
		4 & staff     & NN & friend  & ADJ & 168 & S \\ 
		5 & recommend & VB & high    & ADJ & 161 & Re \\ 
		6 & place     & NN & great   & ADJ & 159 & E \\ 
		7 & go        & VB & back    &  RB & 153 & Rv \\ 
		8 & price     & NN & reason  & ADJ & 150 & P \\ 
		9 & servic    & NN & great   & ADJ & 149 & S \\ 
		10& servic    & NN & friend  & ADJ & 109 & S \\ 
		\hline
\end{tabular}	
\end{table}

We used synonyms and antonyms obtained from an online thesaurus. After the upcycling step, one of the topic domains is newly assigned to 24,379 unstrained bi-terms. Selected examples are presented in \autoref{tab:upcycled}.

\begin{table}[ht]
\centering
\caption{Examples for upcycled uncommon bi-terms \label{tab:upcycled}} 
\begin{tabular}{rllllrl}
  \hline
 & Word 1 & POS 1 & Word 2 & POS 2 & Count & Topic \\ 
  \hline
  1 & ambianc & NN & nice & ADJ &   7 & A \\ 
  2 & cook & VB & delici & ADJ &   7 & F \\
  3 & atmospher & NN & amaz & ADJ &   7 & A \\ 
  4 & bar & NN & small & ADJ &   7 & E \\ 
  5 & beef & NN & order & VB &   7 & F \\ 
  6 & bowl & NN & good & ADJ &   7 & F \\ 
  7 & breakfast & NN & great & ADJ &   7 & F \\ 
  8 & chicken & NN & like & VB &   7 & F \\ 
  9 & choic & NN & excel & ADJ &   7 & Re \\ 
  10 & cocktail & NN & good & ADJ &   7 & F \\ 
   \hline
\end{tabular}
\end{table}

For the TripAdvisor example, we used a general-purpose lexicons data set included in \textbf{R} package `tidytext' \cite{tidytext}. Table \ref{tab:score_biterm} presents scored bi-terms with three categories: positive, negative, and neutral. Note that it is possible to generate not only integer values but also numerical scores.

\begin{table}[ht]
\centering
\caption{Examples of scored clauses through the summarized bi-terms in bold \label{tab:score_biterm}} 
{\footnotesize
\begin{tabular}{rllllr}
  \hline
 & rep.original.review..nrow.final.. & topic & valence \\ 
  \hline
  1 & the prices are fair and the \textbf{food} \textbf{good}.  &  food & 1.00 \\ 
  2 & the \textbf{service} was \textbf{excellent}. & service & 1.00 \\ 
  3 & so if you are in downtown \textbf{area}, this will be a \textbf{good} place ... &  environment & 1.00 \\ 
  4 & the \textbf{sashimi} was so \textbf{fresh} as so reasonable for waikiki. &  food & 1.00 \\ 
  5 & \textbf{crisp} and delicious \textbf{salads}, really tasty garlic prawns... &   food & 1.00 \\ 
  6 & i like \textbf{said} the food was \textbf{excellent}. &  recommend & 1.00 \\ 
  7 & the \textbf{parking} is \textbf{limited} &  environment & -1.00 \\ 
  8 & \textbf{prices} are pretty \textbf{good}. & price & 1.00 \\ 
  9 & \textbf{great} \textbf{place} for hawaiian pizza. &  atmosphere & 1.00 \\ 
  10 & \textbf{price} is \textbf{right} and the service is great. &  price & 1.00 \\ 
   \hline
\end{tabular}
}
\end{table}

\newpage
\section{Conclusion \label{sec6}}

In this paper, we proposed that confirmatory aspect-based opinion mining links online reviews to a vector of pre-specified topics of a specific domain. While the conventional aspect-based opinion mining automatically extracts aspects and then determines the sentiment of reviewers’ opinions, our new algorithm implements a matching process between bi-terms that are linked to clauses and pre-specified topics and then assigns sentiment scores using a lexicon-based sentiment database. 

When analyzing online customer reviews, the proposed method has desirable advantages compared to previous methods, such as Bayesian topic modeling method and conventional aspect-based opinion mining algorithms \cite{Blei03,hu04}. First, there is less ambiguity in topic (or aspect) identification because this algorithm can use a seed of pre-specified topics. Second, the mining results are likely to be more accurate when clause segmentation is successfully implemented during the mining process. Sentence-level algorithms will be less accurate compared to our clause-level-based algorithm because the sentence-level computation may be required to handle more confounding words or expressions for aspect extraction and identification. Third, the proposed algorithm can be easily modified for other mining purposes. Since each clause is summarized to a bi-term and the basic computation units are bi-terms, we can directly incorporate other algorithms to be used on the bi-terms instead of sentences or documents. That is, the proposed algorithm is more compatible with other machine-learning-based algorithms or topic modeling-based approaches.

To implement our confirmatory aspect-based opinion mining, we introduced an algorithm, DiSSBUS, for clause-level summarization that consists of six steps: disintegrating, summarizing, straining, bagging, upcycling, and scoring. The DiSSBUS algorithm simplifies each linguistic computation to connect input text reviews with the pre-specified topics. To test the performance of DiSSBUS, the algorithm was applied to restaurant reviews scraped from TripAdvisor. The topic matching was implemented using eight restaurant attributes as the topics, which include food, service, price, environment, revisit, and recommendation. The results showed that restaurant text reviews were well disintegrated into clauses, and the pre-specified restaurant topics were well assigned to the clauses.

This study is not free from limitations in clause segmentation and sentiment scoring that are commonly found in other topic modeling approaches. First, because the performance depends on the clause segmentation quality, the proposed algorithm may not work well for documents that have extremely lengthy sentences or complex sentence structures. However, our proposed method is valid at least for relatively short text reviews, as we illustrate the matching results from the empirical analysis. Second, scoring is another area that needs to be improved. Scoring is a critical issue not only in the proposed method but also other similar approaches that use sentiment lexicons. The scoring performance often depends on the quality of sentiment lexicons. Because the proposed scoring process is implemented by assigning sentiment scores on the selected bi-terms using the sentiment lexicons, it is also difficult to capture the meaning of longer clauses. To handle these issues, one can incorporate a deep learning algorithm, for example \cite{socher13}, by replacing the current scoring method. This additional work needs to be addressed shortly.

\section*{Acknowledgements}
Jongho Im's research was supported by the Yonsei University Research Fund of 2018-22-0043.

\section*{References}

\bibliographystyle{elsarticle-num} 
\bibliography{CTM_Reference}

\end{document}